# HRI Challenges Influencing Low Usage of Robotic Systems in Disaster Response and Rescue Operations

Shahinul Hoque
Department of Electrical Engineering and Computer Science
*University of Tennessee, Knoxville*
Knoxville, TN
shoque@vols.utk.edu

Farhin Farhad Riya
Department of Electrical Engineering and Computer Science
*University of Tennessee, Knoxville*
Knoxville, TN
friya@vols.utk.edu

Jinyuan Sun
Department of Electrical Engineering and Computer Science
*University of Tennessee, Knoxville*
Knoxville, TN
jysun@utk.edu

*Abstract*—The breakthrough in AI and Machine Learning has brought a new revolution in robotics, resulting in the construction of more sophisticated robotic systems. Not only can these robotic systems benefit all domains, but also can accomplish tasks that seemed to be unimaginable a few years ago. From swarms of autonomous small robots working together to more very heavy and large objects, to seemingly indestructible robots capable of going to the harshest environments, we can see robotic systems designed for every task imaginable. Among them, a key scenario where robotic systems can benefit is in disaster response scenarios and rescue operations. Robotic systems are capable of successfully conducting tasks such as removing heavy materials, utilizing multiple advanced sensors for finding objects of interest, moving through debris and various inhospitable environments, and not the least have flying capabilities. Even with so much potential, we rarely see the utilization of robotic systems in disaster response scenarios and rescue missions. Many factors could be responsible for the low utilization of robotic systems in such scenarios. One of the key factors involve challenges related to Human-Robot Interaction (HRI) issues. Therefore, in this paper, we try to understand the HRI challenges involving the utilization of robotic systems in disaster response and rescue operations. Furthermore, we go through some of the proposed robotic systems designed for disaster response scenarios and identify the HRI challenges of those systems. Finally, we try to address the challenges by introducing ideas from various proposed research works.

## I. Introduction

Compared to the previous robotic systems that used very rudimentary computations, current robotic systems integrate various Machine Learning and Deep Learning algorithms to accomplish sophisticated tasks. Therefore, with the advancement in Machine Learning and Artificial Intelligence domain, we started to see various advanced and powerful robotic systems.

These robotic systems can integrate many sensors to enhance its capabilities. Moving heavy materials, infrared-vision to locate objects more quickly, complex moving mechanisms to make the systems capable of moving in any terrain, robotic systems designed with new and durable materials and alloys capable of withstanding extreme temperatures and pressure, these are some of the capabilities of these new era robotic systems. Therefore, it would be logical to assume that these systems can greatly benefit scenarios like disaster response and rescue missions in harsh environments. However, the reality is that, even with these benefits, the deployment of robotic systems in this sector is still slow. Multiple factors are responsible for the delay of deployment of such robotic systems. One of the factors involves the challenges related to HRI. until now, the majority of the rescue operations in disaster scenarios are conducted by human first responders and hand powered tools as such scenarios require complex and difficult planning and cooperation. To introduce robotic systems in these operations would involve the collaboration between human and robotic systems. Therefore, it is important to first address the Human-Robot interaction challenges to better plan for such scenarios.

In this study, we identify the challenges involving collaboration between humans and robotic systems termed as HRI challenges. We focus our study in HRI challenges involving robotic systems designed for disaster scenarios and rescue missions. We try to categorize these challenges based on environmental challenges, legal challenges, technical challenges, and moral challenges. Then we try to discuss why these challenges are slowing down or hampering the deployment of robotic systems.

Furthermore, we analyze some of the proposed state-of-the-art robotic systems for disaster response scenarios. We try to identify the HRI challenges in those robotic systems. We then discuss how these challenges are obstructing the deployment of robotic systems.

Finally, we try to provide suggestions on how to handle the HRI challenges in each of the categories combining current literature and various proposed ideas. Therefore, the contribution of this paper are as follow-

- Address what are the challenges about Human-Robot Interaction.

- Dividing the challenges into environmental, legal, technical, and moral domains for better understanding.

- Analyze the HRI challenges involving currently proposed robotic systems for Disaster Response and Rescue operations.

- Providing suggestions on how to resolve or lower the impact of these challenges in robotic systems to increase the acceptance of robotic systems in disaster response scenarios.

With a view to achieve our goal, we first discuss what are HRI challenges in general. Then, we focus on the HRI challenges involving disaster response scenarios and rescue operations. Next, we categorize the challenges to various groups. Furthermore, we analyze some of the proposed robotic systems for disaster response and rescue missions and identify the HRI challenges that hamper the mass deployment of the above mentioned robotic systems.
Finally, we provide suggestions on how to mitigate or reduce the impact of the HRI challenges, resulting in the better deployment of robotic systems in disaster response and rescue operations. We also discuss how future robotic systems can be better suited and tested for these scenarios.

## II. HRI Domain

Human-Robot Interaction (HRI) refers to the exchange of information and feedback between a human and a robot agent. Although, there are some common features between HRI and HCI (Human-Computer Interaction), HRI is a totally different domain that involves human interactions with autonomous agents capable of making decisions and taking actions based on environmental feedback. The paper [1], divides using the properties of interaction and the level of autonomy of the robotic agent. The categories are human supervised robots, remote controlled robots, fully autonomous robots, and social robots.

Human supervised robots are robots that are closely supervised by humans. These robots accomplish tasks based on human instructions and need somewhat basic levels of autonomy. A few examples of robotic agents from this category are assembly line robots, delivery robots, and sorting robots in warehouses. All these robots operate in a closed environment and do not need to participate in complex decision making.

Another category of robots are robotic agents controlled from a very long distance. These types of systems need to have some form of advanced autonomy to be able to handle human instructions received periodically. These robotic systems are responsible for route planning, obstacle avoidance, object of interest retrieval and many more types of tasks that require the robotic system to take autonomous decisions as the communication channel used to communicate with the command center might not be good enough to operate the robotic system in real time. Some examples of such robotic systems are rover systems deployed in outer space.

Then there are robotic systems that fully operate autonomously in open environments. Although there are only a few types of agents that can fully operate autonomously in an open-world environment, this is the goal for most robotic systems. Autonomous cars are an example of autonomous robotic systems. Autonomous cars drive in an open environment that is constantly changing. Each agent needs to be able to make complex decisions and path planning. Therefore, these types of systems can be considered as the most advanced form of robotic systems.

We also have a totally different category of robotic systems known as social robots. The main purpose of these types of robots is to provide comfort to elderly people and patients, educational services, and participate in other social interactions with humans. These types of systems might not have complex mechanical functions, but they require the ability to make complex decisions based on the user's response and the environment.

Therefore, it is evident that the HRI is a vast domain containing a wide range of interactions between humans and robotic systems. The interaction has multiple complex factors involving medium of communication, level of autonomy of the robotic system, the operational environment of the system, and interactions among multiple agents. All these types of interactions are studied in the HRI domain to improve future robotic systems and make them more acceptable to society and humans.

## III. HRI Challenges

The apparent complex and boundless interactions between robotic agents and humans result in challenges that need to be addressed for the success of a robotic system. Due to the complex nature of robotic systems, it is very challenging to measure the success rate of a robotic system in any environment [2]. Although a robot designed to identify an object might be considered as a success if it can properly identify the desired object, questions can still remain if another robotic system might be able to accomplish the same task within a shorter time, or in a more power efficient way, or more accurately. The requirements of the application might also influence the importance of certain functions more than other functions. Therefore, though it is hard to design an overall acceptable performing robotic system, it is harder to truly measure the success rate of a robotic system. The task becomes more demanding as human interactions are taken into consideration. Therefore, to be able to truly identify and mitigate the challenges in Human-Robot interactions, the first objective is to identify the challenges in each scenario or application for which a robotic system is designed for. In the next part of the paper, we are going to identify and discuss the specific challenges factoring in the low real-life deployment of robotic systems in disaster response scenarios due to HRI.

### HRI Challenges in Disaster Response

Each Disaster Scenario is unique and a disaster response robotic system not only needs to consider a fully unplanned environment, but also needs to operate in a treacherous environment. Therefore, navigation and path planning is a key challenge in these types of robotic systems [3]. Additionally, as most disaster response robotic systems are designed to be supervised by humans, another challenge that needs to be taken into consideration is the ability of the robotic system to convey information about the environment to the operator so that the operator can be aware of the environment and instruct or operate the robotic system towards a successful mission. We now discuss some of the core HRI challenges that need to be addressed to create more advanced and robust robotic systems.

*A. Environmental challenges*

The complex demands placed upon robotic systems designed for disaster response stem from the diverse nature of their applications. The environmental challenges inherent in disaster areas, characterized by their harsh and perilous conditions, significantly increase the complexity of designing and implementing effective robotic solutions [4]. Navigating through unfriendly terrains, these robotic systems must be remotely operated by humans, necessitating not only adaptability but also an unimpeded ability to move seamlessly amidst the challenges posed by disaster-stricken landscapes. Furthermore, the imperative to furnish real-time visual and sensor data to operators becomes extremely important to ensure timely and informed decision-making during critical situations.

Specific adversities, such as high levels of radiation and contamination, pose challenges, particularly in the area of communication [4]. Establishing robust guidelines or protocols for robotic systems to follow in scenarios of communication interruptions becomes essential, ensuring the continued operation of these systems even in adverse conditions. However, given the uniqueness of each disaster scenario, attempting to comprehensively train rescue teams to operate seamlessly in every conceivable and unimaginable situation proves impractical.

Hence, a lack of confidence in the dependability of robotic systems across diverse environments acts as a deterrent, limiting the integration of more robotic systems in rescue operations. As technology progresses, there is a need for continual refinement and adaptation of these robotic systems to address the ever-evolving challenges posed by diverse and unpredictable disaster scenarios. This underscores the intricate balance between designing versatile and resilient robotic solutions and acknowledging the inherent unpredictability of the environments in which they operate.

*B. Legal Challenges*

Disaster events, characterized by the significant loss of life and property, require the deployment of rescue teams with the primary objective of minimizing such losses. Even in the context of disaster response, rescue teams encounter the complicated realm of tort laws, which describe the consequences of actions resulting in harm or injury to individuals and property [5]. However, the existing legal framework lacks specific provisions addressing tort laws focusing on robotic systems, thereby giving rise to complex legal challenges whenever these systems run afoul of such regulations.

Consequently, rescue teams refrain from employing robotic systems in disaster response and rescue missions to sidestep potential legal liabilities arising from the malfunctioning of such systems. Additionally, the presence of cameras and sensors in robotic systems, capable of capturing and sharing private data during victims' personal and sensitive moments, results in fear regarding potential legal repercussions in the event of data breaches. This concern serves as a deterrent for many organizations considering the utilization of robotic systems in rescue operations.

The absence of established tort laws for robotic systems becomes particularly pronounced when considering the potential dangers associated with their mechanisms when interacting with humans. The complex nature of these legal challenges underscores the vital importance for robotic systems to possess mechanisms capable of swiftly processing human instructions while also incorporating fail-safe systems. These fail-safe measures should be designed with explicit and well-known instructions, facilitating the safe and informed interaction of human operators or any other individuals coming into contact with the robotic agent. As the integration of robotic systems in disaster response becomes increasingly prevalent, the legal landscape must evolve to address these emerging complexities, ensuring a harmonious coexistence between advanced robotic technologies and established legal frameworks. This calls for a proactive approach in shaping legislation that adequately encompasses the unique challenges posed by robotic systems in disaster scenarios, fostering a comprehensive legal framework that prioritizes safety, accountability, and responsible use.

*C. technical challenges*

A common goal of engineers has always been to design a human-like robotic system that can mimic a human and take part in the same functions that a human can do. Known as a humanoid, a human-like robot would be the ideal robotic system to participate in disaster response scenarios as it would be able to accomplish all the tasks that a rescue team member can do with the added benefit of being disposable or preventing further loss of life. Furthermore, such a robotic system could participate in human rescue teams and fully understand and cooperate with human team members in rescue operations. However, there are many technical limitations that restrict the design of such a robotic system [6]. Initially, even if it was feasible to create hardware capable of mimicking human movements, such a robotic system would rely on vocal communication and speech when interacting with humans. Nevertheless, given that robotic systems inherently understand only binary language, any human command must undergo conversion into binary functions for execution. This process requires assigning a precise function to each human instruction. Moreover, the subjective nature of sentence meanings, influenced by the context and various factors, poses technical challenges in developing a robotic system that can accurately associate each sentence with the appropriate subjective interpretation.

Furthermore, despite possessing a degree of autonomy, robotic systems inherently rely on human instructions to navigate and operate effectively in unsuitable environments. However, this interdependence introduces a critical prerequisite – the operator's comprehensive understanding of the technical capacity and capabilities of the robotic system. The efficacy of human-robot collaboration hinges on the operator's ability to provide instructions aligned with the capabilities inherent to the robotic system in use.

In the context of disaster scenarios, human team members naturally possess an inherent understanding of each other's capabilities, fostering seamless cooperation without the apprehension of malfunctions or errors. This interpersonal familiarity allows for effective coordination and adaptability in dynamic situations. Conversely, the dynamics shift significantly when it comes to robotic systems. Unlike human counterparts, robotic entities lack the intuitive understanding that humans have about each other's capacities, introducing a distinct challenge.

The absence of innate comprehension between humans and robotic systems necessitates meticulous training and communication protocols. Operators must possess not only a deep knowledge of the robotic system's technical nuances but also a keen awareness of its limitations. This awareness is crucial for issuing instructions that optimize the robotic system's performance while mitigating the risk of errors or malfunctions. Therefore, as the integration of robotic technologies continues to advance, emphasis on comprehensive operator training and clear communication channels becomes pivotal to unlocking the full potential of these systems in disaster response scenarios.

*D. moral challenges*

In a disaster scenario, the main objective of a rescue team is to mitigate the loss of life and rescue stuck victims. However, there are situations where rescue missions become recovery missions, and euthanizing actions need to be performed. In those types of situations, the moral consequences of a human team member performing the euthanizing action is not frowned upon as a human has better judgment of good and bad. It is very challenging to embed these judgemental values into the programming of an autonomous robotic system. Situations where it is important to sacrifice something for the greater good, is not something an autonomous robotic system understands [8].

Additionally, societal sentiments concerning the treatment of deceased bodies by robotic systems reflect a prevailing perception of insufficiency in terms of respect, compared to the way humans are expected to handle such sensitive tasks. Consequently, when faced with situations demanding the utilization of robotic systems for the operations involving deceased bodies, rescue teams find themselves subjected to disapproval and criticism from both the general public and various organizations. This societal disapproval becomes a significant factor influencing the decision to reduce the incorporation of robotic systems in missions or operations that involve the handling of deceased bodies. This reluctance, in turn, results in missed opportunities to harness the benefits that robotic systems could offer in such scenarios, underscoring the complex cooperation between societal expectations and the broader adoption of technological solutions in sensitive domains.

IV. HRI CHALLENGES WITH MODERN ROBOTIC SYSTEMS

In recent times, many research groups have proposed various robotic systems that seem to be able to handle and successfully demonstrate their capabilities in various scenarios. One of the most popular challenges is The Grand Challenge by DARPA [9]. In this challenge, various research groups present their autonomous vehicle capable of running a long race through tough terrain and challenging environments. The following figure demonstrates such a vehicle.

Despite the capability for autonomous driving, vehicles in disaster scenarios require the ability to engage with human operators or drivers for guidance. Additionally, the absence of information regarding path decisions poses challenges for both technical and environmental considerations. In such instances, a human operator may struggle to anticipate the agent's next move, emphasizing the importance of effective communication and shared decision-making between autonomous systems and human counterparts. These challenges underscore the need for robust systems that seamlessly integrate autonomous capabilities with human

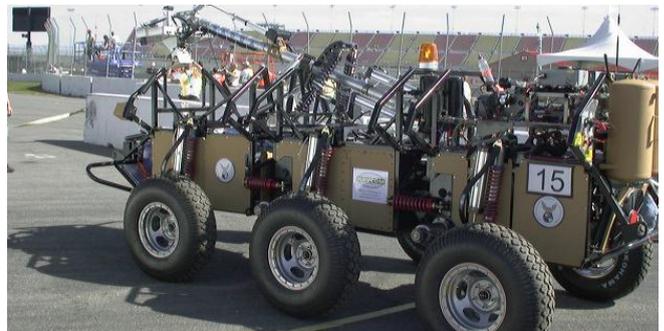

Figure 1: Autonomous vehicle designed for the DARPA Grand Challenge 2004. Source: DARPA.

Robotic systems designed for disaster response and rescue operations during earthquakes might be beneficial in rescue as elapsed time is very important in an earthquake disaster. In the past, there were many situations where victims were stuck for more than a few days under collapsed buildings. During these types of disasters it is important to quickly locate and rescue victims stuck under collapsed structures to lower the number of lost life [10]. However, the ability of robotic systems like NIFTi UGV are designed to identify and locate victims, but lack the ability to actively rescue victims due to technical limitations. Furthermore, there are many scenarios where victims stuck under debris and rubble need moral support before rescue operations can take place. A robotic system only designed to locate victims would not be able to provide that support due to the complexity to integrate moral and social values into a robot due to technical and moral challenges.

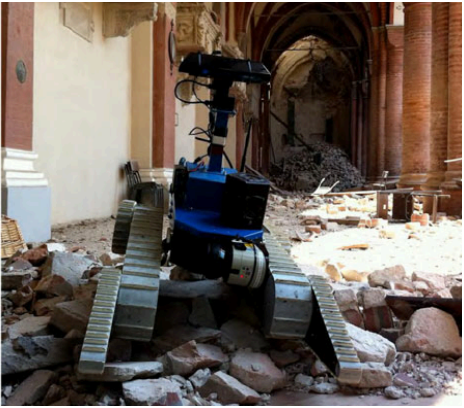

Figure 2: NIFTi UGV robot designed for rescue missions in earthquake affected disasters.

## V. MITIGATING HRI CHALLENGES

Although the HRI or Human-Robot Interaction field is still new and needs more research to find out the key challenges, there are ways to limit or mitigate the current obstacles that deter the deployment of robotic systems.

### A. User-Centered Design

Initiating an effective approach to integrating robotic systems into disaster response begins with the foundational step of designing these systems according to the specific needs of rescue team members and the essential requirements of the victims. This entails a meticulous analysis of the challenges and demands faced by rescue teams in diverse disaster scenarios. Understanding the unique operational environments, constraints, and tasks encountered during rescue operations is crucial to tailor the robotic systems accordingly.

Moreover, a user-centered design approach is essential, emphasizing collaboration with rescue team members to incorporate their insights and experiences into the development process. This ensures that the robotic systems are not only technologically advanced but also aligned with the practical requirements and preferences of those who will be actively using them in the field.

Considering the needs of victims is equally vital in this design phase. Robotic systems should be equipped with features that enable them to address the diverse needs and vulnerabilities of individuals affected by disasters. This could involve functionalities related to medical assistance, communication, or the retrieval of critical resources in a manner that ensures the safety and well-being of the victims.

In essence, the first step in enhancing the role of robotic systems in disaster response is a conscientious and collaborative design process. By placing a strong emphasis on understanding and incorporating the perspectives of both rescue team members and victims, these robotic systems can be tailored to serve as valuable assets in the intricate and challenging landscape of disaster scenarios.w

### B. Natural Language Processing

The emergence of AI tools, particularly those leveraging large natural language models, has ushered in a new era where machines can adeptly mimic human speech. This technological advancement opens up a range of possibilities, and one notable application is the potential integration of these AI tools to provide moral support to victims in challenging situations.

These AI-driven systems, equipped with sophisticated natural language processing capabilities, can emulate human-like interactions and expressions. In disaster scenarios where victims often face emotional distress and uncertainty, the deployment of AI tools for moral support becomes a compelling proposition. These tools can offer empathetic responses, provide reassurance, and offer a virtual presence that may alleviate feelings of isolation and fear among victims.

The key advantage lies in the ability of AI tools to continuously improve and adapt their responses based on a vast amount of linguistic data they are trained on. This enables them to tailor their interactions to the specific emotional needs and nuances of individuals in distress. Additionally, these AI-driven systems can operate tirelessly, ensuring a consistent and always-available source of support.

However, ethical considerations surrounding the use of AI in emotionally sensitive contexts should be carefully addressed. Safeguarding privacy, ensuring transparency in AI interactions, and respecting cultural differences are crucial aspects of responsible deployment. Striking the right balance between the technological capabilities of AI and the human touch required for genuine emotional support is a delicate yet essential challenge in leveraging these tools effectively.

In summary, the advent of AI tools capable of mimicking human speech presents a promising avenue for integrating technology into disaster response efforts. By providing virtual moral support, these systems have the potential to complement human efforts and enhance the overall well-being of victims facing difficult situations.

### C. Training

Indeed, addressing both environmental and technical challenges in the context of disaster response can significantly benefit from extensive training simulations that accurately replicate the diverse and complex conditions of disaster environments. While there are established training simulations for nuclear disaster scenarios, there is a noticeable gap when it comes to preparing for more prevalent and varied disasters such as wildfires, tornadoes, floods, earthquakes, and blizzards.

Extended and realistic training simulations play a pivotal role in preparing response teams for the intricacies of each disaster type. For instance, simulations of wildfires can expose teams to the rapid and unpredictable spread of fires, requiring effective coordination and decision-making. Simulating tornado scenarios can help responders understand the dynamics of sudden and powerful wind events, necessitating swift and strategic actions to ensure public safety.

Training in flood scenarios can involve simulations of rising water levels, emphasizing the challenges of evacuation, rescue, and relief operations. Earthquake simulations can replicate the chaotic aftermath, testing teams on their ability to navigate unstable environments and provide immediate assistance. Similarly, blizzard simulations can prepare responders for the unique challenges posed by extreme cold, heavy snowfall, and disrupted transportation systems.

By expanding the scope of training simulations to encompass a broader range of disaster scenarios, response teams can enhance their adaptability and readiness. These simulations should not only replicate the environmental conditions but also integrate technological challenges, such as the integration of robotic systems and AI-driven tools, to ensure a comprehensive training experience.

In addressing the current gap in training for more common disasters, there is an opportunity to leverage advanced technologies, including virtual reality (VR) and augmented reality (AR), to create immersive and dynamic training environments. These technologies can provide realistic scenarios that closely mimic the challenges responders would face in real-world disaster situations, offering a more effective and versatile training experience.

In conclusion, expanding training simulations to cover a wider array of disaster scenarios, coupled with the integration of cutting-edge technologies, can significantly contribute to mitigating both environmental and technical challenges in disaster response efforts. This approach ensures that response teams are well-prepared and equipped to handle the complexities of the disasters they may encounter in the diverse landscapes they serve.

*D. Emotional Recognition*

While integrating emotional values into robotic systems poses challenges, it is crucial to encourage more studies and research projects aimed at simulating diverse approaches. The objective is to enhance the incorporation of emotional properties into robotic systems specifically designed for human rescue operations. This endeavor seeks to explore various options and methods that can effectively embed emotional capabilities, thereby enabling these robotic systems to offer more meaningful moral support in challenging situations. By fostering ongoing research in this area, we can advance the development of emotionally intelligent robotic systems that better cater to the emotional needs of individuals in rescue scenarios.

*E. Legal Guidelines*

It's important for the government to make clear rules about how we can use robots in rescue missions. Also, the people who make these rules should join in discussions and meetings to learn more about how robots are used in emergencies. By working together, they can make rules that keep everyone safe and make sure robots are used in the right way during disasters. This way, we can use robots to help us during tough times while making sure we follow the rules and keep things fair for everyone.

VI. CONCLUSION

The clear advantages of using robotic systems in rescue operation and disaster response scenarios clearly overshadows the disadvantages. Even then, the challenges speciality involving human-robot interaction limit the mass deployment of robotic systems in such circumstances. To tackle the challenges at hand and identify effective solutions, it is essential to conduct additional research and engage in public discussions. By delving deeper into these issues through research endeavors, we can gain a better understanding of the complexities involved. Furthermore, fostering public discussions ensures a diverse range of perspectives are considered, promoting collaborative efforts to address challenges and devise comprehensive solutions. This combined approach of research and public discourse contributes to a more informed and inclusive problem-solving process.

Successfully addressing and mitigating these challenges holds the key to developing the next generation of robotic systems. These advanced systems have the potential to offer enhanced assistance and support in disaster and rescue missions. By overcoming the hurdles posed by environmental, technical, and societal factors, we pave the way for the creation of more robust, adaptable, and effective robotic technologies. This, in turn, will significantly improve the capabilities of these systems to provide valuable aid and support in times of crisis, marking a substantial step forward in the evolution of disaster response technology.

REFERENCES

1. Sheridan, T. B. (2016). Human–Robot Interaction: Status and Challenges. Human Factors, 58(4), 525-532. https://doi.org/10.1177/0018720816644364

2. Z. Z. Bien and D. Stefanov, Advances in Rehabilitation Robotics: Human-friendly Technologies on Movement Assistance and Restoration for People with Disabilities. Berlin Heidelberg, Germany: Springer, 2004.

3. W. He, Z. Li and C. L. P. Chen, "A survey of human-centered intelligent robots: issues and challenges," in IEEE/CAA Journal of Automatica Sinica, vol. 4, no. 4, pp. 602-609, 2017, doi: 10.1109/JAS.2017.7510604.

4. Smith, R.; Cucco, E.; Fairbairn, C. Robotic Development for the Nuclear Environment: Challenges and Strategy. Robotics 2020, 9, 94. https://doi.org/10.3390/robotics9040094

5. Guerra, A., Parisi, F., & Pi, D. (2022). Liability for robots I: Legal challenges. Journal of Institutional Economics, 18(3), 331-343. doi:10.1017/S1744137421000825

6. Dario, P., Guglielmelli, E. and Laschi, C. (2001), Humanoids and personal robots: Design and experiments. J. Robotic Syst., 18: 673-690. https://doi.org/10.1002/rob.8106

7. Human-in-the-loop Control of a Humanoid Robot for Disaster Response: A Report from the DARPA Robotics Challenge Trials

8. Aaron M. Johnson aaronjohnson@alumni.cmu.edu & Sidney Axinn axinn@temple.edu (2013) THE MORALITY OF AUTONOMOUS ROBOTS, Journal of Military Ethics, 12:2, 129-141, DOI: 10.1080/15027570.2013.818399

9. DARPA. (n.d.). *The Grand Challenge*. DARPA RSS. https://www.darpa.mil/about-us/timeline/-grand-challenge-for-autonomous-vehicles

10. G. -J. M. Kruijff et al., "Rescue robots at earthquake-hit Mirandola, Italy: A field report," 2012 IEEE International Symposium on Safety, Security, and Rescue Robotics (SSRR), College Station, TX, USA, 2012, pp. 1-8, doi: 10.1109/SSRR.2012.6523866.